# A Hybrid Method for Traffic Flow Forecasting Using Multimodal Deep Learning

**Shengdong Du[1], Tianrui Li[1,*], Xun Gong[1] and Shi-Jinn Horng[2,*]**

[1]School of Information Science and Technology, National Engineering Laboratory of Integrated Transportation Big Data Application Technology, Southwest Jiaotong University, Chengdu 611756, China

[2]Department of Computer Science and Information Engineering, National Taiwan University of Science and Technology, Taipei 10607, Taiwan

*Abstract*

Traffic flow forecasting has been regarded as a key problem of intelligent transport systems. In this work, we propose a hybrid multimodal deep learning method for short-term traffic flow forecasting, which can jointly and adaptively learn the spatial-temporal correlation features and long temporal interdependence of multi-modality traffic data by an attention auxiliary multimodal deep learning architecture. According to the highly nonlinear characteristics of multi-modality traffic data, the base module of our method consists of one-dimensional Convolutional Neural Networks (1D CNN) and Gated Recurrent Units (GRU) with the attention mechanism. The former is to capture the local trend features and the latter is to capture the long temporal dependencies. Then, we design a hybrid multimodal deep learning framework for fusing share representation features of different modality traffic data by multiple CNN-GRU-Attention modules. The experimental results indicate that the proposed multimodal deep learning model is capable of dealing with complex nonlinear urban traffic flow forecasting with satisfying accuracy and effectiveness.

## 1. Introduction

With the rapid growth of vehicles and the progress of urbanization, the annual cost of traffic jams in urban cities is increasing rapidly, which causes the low efficiency of transportation networks, and results in the loss of time, waste of fuel and excessive air pollution. Therefore, research on the forecasting of urban traffic flow is crucial and it has been regarded as a key problem of intelligent transport management [1], which is also an important means to guide the scientific decision-making of traffic management. Early diagnosis of congestion occurrence and forecast traffic flow evolution are considered to be a key measure to determine traffic bottlenecks, which can be used to support intelligent transport management in the auxiliary.

Over the last decades, many researchers have made efforts to traffic congestion diagnosing and traffic flow forecasting [2-5]. However, most of these studies rely on mathematical equations or simulation techniques to describe the evolution of network congestion. People, weather, accidents and other factors are then usually involved in the transportation network and it is difficult to accurately represent them and study in mathematical models. These traditional methods include classical shallow learning algorithms, such as autoregressive statistics for time series (ARIMA) [4], artificial neural networks (ANNs) [7, 9], and Support Vector Regression (SVR) [6], etc. With the development of the Internet of Things and the application of traffic sensor data acquisition technology, the era of traffic big data has arrived. Traffic congestion forecasting is increasingly dependent on a variety of sensors and related data acquisition equipment to collect the relevant data, such as traffic, speed, journey time, density, weather and accidents data, etc. However, the above traditional models cannot adapt well to the new conditions. Thus, traffic flow forecasting requires data-driven model support [8].

The most representative data-driven model is deep learning [10], which can automatically extract the relevant deep features of traffic data from multiple levels. Recently, deep learning has proven to be very successful in many areas, e.g., image, audio and natural language processing tasks since the breakthrough of Hinton et al. [11], and these researches

This work was partially supported by the National Natural Science Foundation of China (No. 61773324, 61603313), and the Fundamental Research Funds for the Central Universities (No. 2682017CX097), and MOST under 106-2221-E-011-149-MY2 and 107-2218-E-011-008.

show that deep learning models have a superior or comparable performance with state-of-the-art methods in many fields [12-15,19,26]. Because traffic congestion process and traffic flow evolution are dynamic and nonlinear in nature, and deep learning model can learn the deep features of traffic data without prior knowledge. For traffic flow forecasting, the deep learning method has drawn a lot of research interest [16, 17]. For instance, Lv et al. presented a novel deep-learning-based traffic flow prediction method, which used stacked auto encoder model to learn traffic flow features and the spatial-temporal correlations inherently [16].

Multimodal deep learning involves the fusion learning of multiple data sources [21, 22]. For example, Karpathy et al. proposed a multimodal fusion method of image and text data which is based on LSTM and CNN to complete the image description task. The input part uses CNN to extract the characteristics of the image, and the output part uses LSTM to generate the text [13]. In addition, the encoder-decoder (also called sequence-to-sequence) deep learning model has attracted a lot of attention from researchers as a simplified and automatic method for sequence data processing [33]. Especially, the attention mechanism has been widely applied to natural language and speech processing tasks [34].

However, to the best of our knowledge, only a little research has been conducted to combine multimodal deep learning and attention mechanism for traffic sequence data analysis. Therefore, it is highly desired to develop a multimodal deep learning framework to model traffic flow evolution. This paper aims at developing a data-driven traffic flow forecasting paradigm which is based on multimodal deep learning theory. An end-to-end and adaptively multimodal deep learning model is presented to solve the traffic flow forecasting problems which based on CNN-GRU basic modules with attention mechanism supported. The main contributions of this paper are summarized as follows:

● Traffic flow forecasting is challenging under non-free-flow situations because the traffic data (such as flow, speed, density, journey time etc.) are always highly nonlinear and non-stationary, which are affected by different components under different traffic conditions (e.g., peak hours, weather, incidents etc.). We firstly proposed a end-to-end multimodal traffic related sequential data processing framework which focuses on the impact on local spatial features and long dependency features and spatial-temporal correlations via adaptively multimodal deep learning model.

● As the basic module of our adaptively and jointly multimodal deep learning model, a CNN-GRU based and attention mechanism-supported hybrid structure is proposed to solve the traffic flow forecasting problems, which can learn the long temporal dependencies and spatial-temporal correlation features of each traffic related sequence data. This work aims to improve the multi-level feature learning ability by using a multimodal deep learning architecture, which is crucial to make it more robust and flexible in handling traffic flow forecasting problems.

● We demonstrate the effectiveness of our model by testing it on real traffic flow datasets, and the experimental results indicate that our model has good forecasting performance and generalization ability. It is also presented that the proposed multimodal deep learning model (especially with CNN-GRU-Attention module) has better prediction ability than typical shallow learning and baseline deep learning models.

The structure of this paper is described as follows: Section II presents the related work. Section III summarizes the main characteristics of the multimodal deep learning method. The motivation and design of the proposed multimodal architecture for traffic flow forecasting are also addressed, including how to extend and integrate the basic deep learning models to the proposed novel framework. Section IV presents empirical studies using real traffic dataset and evaluates the performance of the proposed method. The last section offers conclusions and directions for future research.

## 2. Related Work

Traffic flow forecasting has a good study history in the transportation literature. A large number of traffic flow prediction methods have been developed to help effective management and decision of intelligent transportation systems [1, 2, 32, 35]. Williams et al. used ARIMA to modeling and forecasting vehicular traffic flow [4]. Castro-Neto et al. proposed an online learning short-term traffic flow forecasting method which is based on the SVR model under typical traffic conditions [6]. Lippi et al. reviewed existing approaches to short-term traffic flow prediction under the common view of probabilistic graphical models [2]. Chan et al. presented an optimized ANN model short-term traffic flow forecasting using a hybrid exponential smoothing and Levenberg–Marquardt method [7]. Sun et al. proposed a Bayesian network-based approach for short-term traffic flow forecasting [5].

Recently, deep learning also has been widely applied to traffic pattern recognition and traffic flow forecasting [16-18, 31]. Song et al. proposed a framework which is based on deep learning models that aim to understand human mobility and transportation patterns from big heterogeneous data [25]. Yang et al. developed an optimized structure of the traffic flow forecasting model which is based on stacked auto-encoder deep learning approach [18]. Huang et al. applied the deep learning approach to transportation research, which incorporates multitask learning (MTL) in the deep architecture, and exhibits its superiority of predicting traffic flow over traditional methods [17]. Also, the deep convolution network could extract patterns, e.g., citywide crowd congestion and Zhang et al. used deep residuals network to learn how the congestion is evolving [24]. Moreover, the encoder-decoder deep learning model with attention mechanism has become a hot research topic [33]. For example, Chorowski et al. extended the attention-mechanism with features needed for speech recognition, and experimental results showed that the proposed model reaches a competitive performance [34].

Multimodal deep learning [21], which combines the strengths of various deep learning models (especially LSTM and CNN), has received an increasing interest in the computer vision domain, e.g., image captioning and image classification. Many related types of research have studied based on multimodal deep learning, which is often effective for improving the prediction performance of deep learning methods [13, 22]. Srivastava et al. used Deep Boltzmann Machine for learning a multimodal model which consists of multiple and diverse input modalities. And it is useful for classification and information retrieval tasks [23]. On the other hand, multimodal deep learning can combine the strengths of multi-modality data sources (e.g., traffic flow, speed, dense, journey time, weather and accidents data, etc.), However, such multimodal deep learning methods have not been well studied for traffic flow research.

Our methodology provides an alternative to previously proposed traffic flow forecasting methods and traditional shallow learning models for traffic congestion analysis. The proposed method is not only to capture nonlinear spatial-temporal effects of both local trends and long dependences on single modality traffic data, but also make use of the multi-modality traffic data (e.g., traffic speed, traffic flow, weather, accidents and traffic journey times, etc.) by multimodal deep fusion learning. Moreover, our method focuses on the impact on local spatial and long temporal features and discovers spatial-temporal correlations between speed-flow-journey time in multi-modality traffic data.

## 3. Methodology

### 3.1 Problem definition and motivation

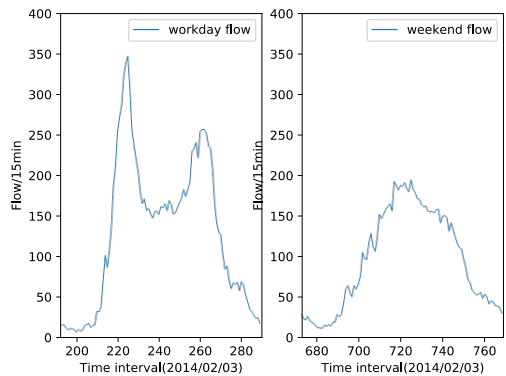

Fig. 1. Workday vs. Weekend traffic flow, from Site A414 between M1 J7 and A405 link road of Highways Agency in England [30].

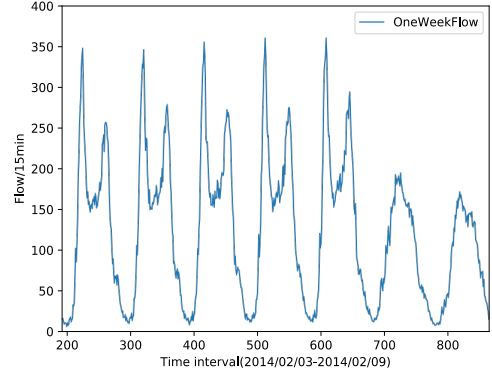

Fig. 2. One week traffic flow data, from Site A414 between M1 J7 and A405 link road of Highways Agency in England [30].

Traffic flow forecasting is a challenging problem in the field of intelligent traffic management. Its goal is to anticipate changes in the number of vehicles at observation points (e.g., crossroads or stations) over time. The time interval is usually set to 5, 15 or 30 minutes. Typical traffic flow data are shown in Fig. 1 and Fig. 2.

In general, traffic flow is an average number of vehicles observed for the link in a given time period. Here we use $f_{i,t}$ indicates the average number of vehicles passing through the $i$th observation point (such as road junction or station) during the $t$th time interval. Traffic flow forecasting problem is: At time $T$, the traffic flow forecasting mission is to predict the traffic flow $f_{i,T+1}$ at time $T + 1$ or $f_{i,T+n}$ at time $T + n$ based on the history traffic data (e.g., traffic flow sequence $F = \{f_{i,t} | i \in O, t \in \{1,2,...,T\}\}$, where $O$ is the set of all observation points).

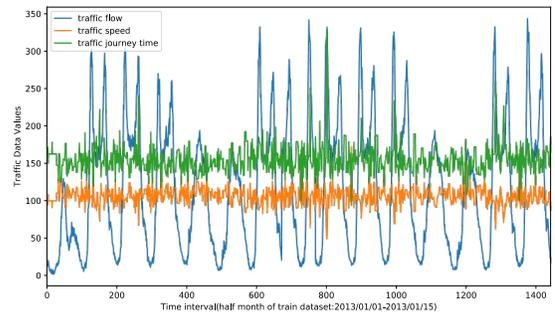

Fig. 3. The interdependences and correlations of multimodal traffic data (take traffic flow, speed and journey time data as an example).

There are two key problems with data-driven traffic flow forecasting method:

● How to deal with the characteristics of the spatial-temporal features of single modality data is the first key point. Take the traffic flow data itself as an example (See one-week data points in Fig. 2, 2014/02/03-2014/02/09). Because the sequence of historical local spatial feature describes the contextual information of traffic flow time series, it naturally affects the evolution of the following trend (we call it a local trend. There is a correlation between local neighbors and these points tend to act similarly). That is to say, the nearby data points and periodic interval data of traffic flow typically have a strong relationship with each other. There have correlations between local neighbor features and long dependency features of traffic flow time series data.

● How to deal with the interdependence of multi-modality traffic data is the second key problem. Traffic flow data have sharp nonlinearities resulting from transitions from free flow to break down and then to congested flow. Traffic flow forecasting is challenging under non-free-flow situations (e.g., peak hours, incidents, work zones, etc.) due to rapidly changing traffic conditions. It is related to many factors, e.g., traffic speed (the average speed of vehicles entering the junction to junction link within a given time period), traffic journey time (the average pass time of vehicles to travel across the link road, also called pass time), traffic accidents or weather conditions, etc. Those influences are complex and highly non-linear and it is hard to precise forecast traffic flow for a specific time and place because these factors are inherently interdependent (See Fig. 3).

### 3.2 Overview of the multimodal deep learning framework

To tackle the above problems, we propose a hybrid traffic flow forecasting method by using multimodal deep learning models. Generally speaking, it is difficult to use a shallow model for fusion modeling due to the different statistical characteristics of different modality data (each modality having different representation and correlational structure). Multimodal deep feature learning refers to fusion at the multi-feature level, such as feature concatenation or a linear combination of local trend features and long dependency features of multi-modality traffic data.

In the following section, we describe the hybrid multimodal deep learning framework for traffic flow forecasting (HMDLF for short). It is motivated by the combination of CNN and GRU neural networks with attention mechanism as a CNN-GRU-Attention module, which considers the spatial-temporal dependency features of single modality traffic data. We then fuse share representation features of different modality traffic data based on multiple CNN-GRU-Attention modules. Fig. 4 is the graphical illustration of our proposed multimodal deep learning framework.

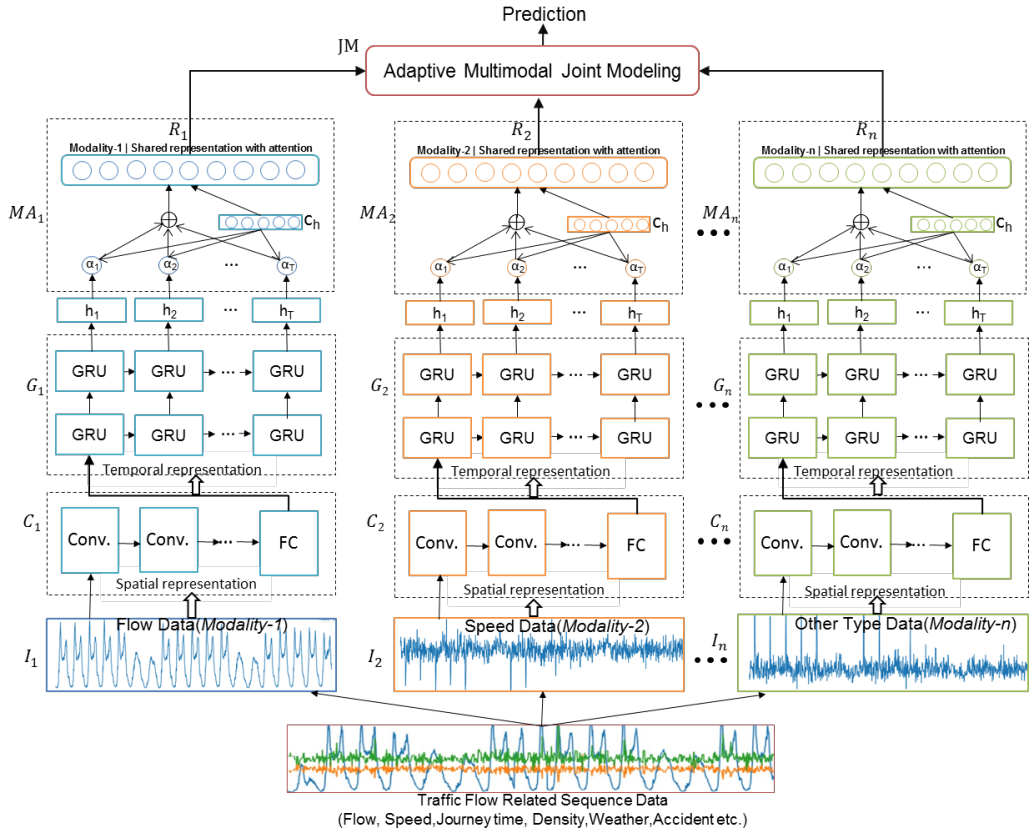

Fig. 4. A hybrid multimodal deep learning framework for traffic flow forecasting diagram (HMDLF for short). Schematic illustration of the proposed method in hierarchical feature representation and multimodal fusion with deep learning for traffic flow forecasting.

It can be found from Fig. 4 that the overall framework consists of three components: convolution model (1D CNN) for spatial representation learning of the local trends of sequence data; GRU models for temporal representation learning of the long dependency features; Attention mechanism for important features learning and the final adaptively joint model for multimodal data representation fused learning.

In representing the different traffic data modality as spatial-temporal features, the first step is to train CNN and GRU models to extract deep correlation features. For a given training modality dataset $I_i$, each spatial-temporal feature level pair can be represented as follows:

$$C(I_i) \rightarrow S_i \quad G(S_i) \rightarrow S_i T_i \quad MA(S_i T_i) \rightarrow R_i \quad (1)$$

Here $S_i$ and $T_i$ denote spatial and temporal correlation features that will be extracted from each traffic modality input dataset $I_i$ with CNN model $C$ and GRU model $G$, respectively. $MA$ represents the multimodal feature level fusion layers with attention mechanism ($R_i$ is the shared representation of $S_i$ and $T_i$ with attention assisted learning). To learn multimodal representation of different modality traffic data (e.g., speed, flow, journey time, weather, etc.), we design a joint and adaptive deep learning framework for fusing spatial-temporal shared features of different modality traffic data. The multimodal joint model is then described as:

$$JM((R_1, R_2, \ldots, R_n), W^i, b^i) \rightarrow \pi, \quad i = 1,2,\ldots n \quad (2)$$

The $\pi$ denotes the joint fusion representation for different learned spatial-temporal feature pair $R_i$ extracted from multi-modality datasets. $W^i$ and $b^i$ are weights and biases that will be learned by the joint model with multimodal training datasets. $i$ denotes each modality input. The training objective function of HMDLF model is described as follows:

$$\operatorname*{argmin}_{\theta} C_i = \frac{1}{n}\sum_{i=1}^{n}\sum_{j=1}^{m} ||\hat{y}_i^j - y_i^j||^2 + \frac{\lambda}{2}\sum_{l} ||W_l^i||_F^2 \quad (3)$$

The final model training problem is to minimize overall error $C_i$ of training samples for each modality, where $i$ denotes each modality input ($i$=1, 2, ..., $n$), $j$ indicates the input sample number of a single modality data ($j$=1, 2, ..., $m$) and $W_l^i$ represents the weight parameter of each layer $l$ of $i$-th modality input. $\theta$ is the parameter space including $W_l^i$ and $b_l^i$ of each layer and $\lambda$ controls the importance of the penalty or regularization term of the objective function.

Through the above process, 1D-CNN is used to capture the local trend features of the special modality sequential data, and GRU with Attention layer are utilized to learn features of both short-term time variation and long-term dependency periodicity. Then we share those spatial-temporal features into a feature-level based fusion layer. For multimodality traffic data, the processing method is the same as that of the single modality data processing. The difference is that the different modality data sharing representation features are combined by multimodal Joint Model. After that, we feed those joint fusion features into a regression layer for final prediction.

### 3.3 CNN for spatial local trends learning

CNN is a feed-forward neural network. Its artificial neurons can respond to the part of the coverage area. CNN has good image processing performance, and some researchers also used it for time series analysis. A classical CNN has three cascaded layers (e.g., convolutional, activation and pooling layers) [12]. Due to the time shift and periodicity of traffic flow data, we use one-dimensional CNN to carry out the sequence local trend learning, which extracts the local trend features by convolution operations of CNN, and these features can be served as more deep representation in the proposed multimodal deep learning framework. The CNN's three layers are described as follows:

$$c_j^l = \sum_i x_i^{l-1} * w_{ij}^l + b_j^l \quad (4)$$
$$x_j^l = \emptyset(c_j^l) \quad (5)$$
$$x_j^{l+1} = pool(x_j^l) \quad (6)$$

Here, Equation (4) represents the convolutional operation, Equation (5) represents the activation function and Equation (6) indicates the function used for pooling. $x_i^{l-1}$ and $c_j^l$ represent the input and output of convolution layer, respectively, where $l$ represents the involved layer. $c_j^l$ and $x_j^l$ denote the input and output of activation layer, respectively.

### 3.4 CNN-GRU module (with attention mechanism) for long temporal dependencies and spatial-temporal correlation features learning

Recurrent neural network (RNN) is a popular dynamic system for handling sequence tasks, e.g., time series prediction or speech recognition. The structure of an RNN enables it to maintain a state vector that implicitly contains historical information about all the past elements of a sequence. An RNN unfolds in time and can be considered as a deep neural network with an infinite number of layers:

$$s_t = f(Ux_t + Ws_{t-1}) \quad (7)$$
$$y = g(Vs_t) \quad (8)$$

In above equations, $x_t$ represents the input information at time $t$. $s_t$ represents the hidden state at time $t$, which indicates the memory of RNN network. $f$ represents the activation function, such as *tanh* or *ReLu* function. $g$

represents the activation function for the output layer (e.g., *softmax*) and $y$ indicates the output of the network at time $t$. Unlike CNN, a RNN shares the same parameters across all time steps, and $U$, $V$, $W$ indicates the shared network parameters.

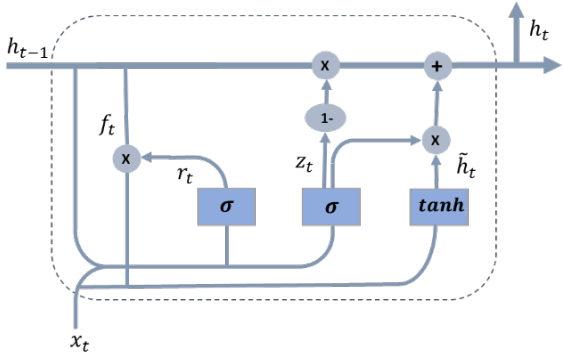

Fig. 5. A typical GRU block diagram, visualization idea by Christopher Olah [29].

Based on variants of RNN, Long Short-term Memory network (LSTM) can be used to capture the long dependency features of sequence data which is proposed by Hochreiter and Schmidhuber [20]. It is very popular and capable of processing sequence learning tasks. A well-known variant network based on LSTM is the Gated Recurrent Unit (GRU), which is proposed by Cho et al. [14] (see Fig. 5). It combines the forget gates with the input gates to a single update gate. The GRU model is simpler and has fewer parameters than the LSTM, and has been shown to outperform or keep the same performance as LSTM on some tasks. Therefore, instead of using LSTM, we use GRU for multimodal long dependency features learning in the proposed framework.

Fig. 5 is a typical GRU block diagram. The memory cell of each GRU contains four main components. These gates allow cells to save and access information for a long period of time. The long temporal dependencies learning block GRU calculates the hidden states by a set of equations listed as follows:

$$z_t = \sigma(W^{(z)} \cdot [h_{t-1}, x_t]) \quad (9)$$
$$r_t = \sigma(W^{(r)} \cdot [h_{t-1}, x_t]) \quad (10)$$
$$\tilde{h}_t = tanh(W \cdot [r_t * h_{t-1}, x_t]) \quad (11)$$
$$h_t = (1 - z_t) * h_{t-1} + z_t * \tilde{h}_t \quad (12)$$

In these equations $z_t$, $r_t$ are related to the update gate and reset gate, respectively. The update gate $z_t$ decides how much the unit updates its activation, and $\sigma$ is the activation function. Similarly, the reset gate $r_t$ allows the unit to forget the past information (when $r_t$ equals 0). The candidate activation $\tilde{h}_t$ is computed with the reset gate $r_t$ (which decides if forgets the past) and $*$ denotes an elementwise multiplication. Finally, $h_t$ represents the actual activation of the proposed GRU unit at time $t$, which is a linear interpolation between the previous activation $h_{t-1}$ and the candidate activation $\tilde{h}_t$. As the above process, each hidden unit of the GRU has a reset gate and an update gate, which is learning to capture dependencies over different timescales. If the reset gate is activated, then it tends to learn short-term dependencies; otherwise, it tends to catch long-term dependencies. But there is a problem in the learning process, it is due to the fact that the spatial-temporal dependency context (observation fragment) does not contribute equally to the deep representation of a time series sequence. In order to solve this problem, this paper proposes the attention mechanism for the hybrid traffic flow forecasting model. Here the attention layer selects spatial-temporal context in the shared representation of each modality data to which HMDLF model should attend, and promotes to predict the next time series value precisely. It constructs attention context vectors for different time step prediction as a weighted sum of the hidden states of the GRU layer output, which is described as follows:

$$e_t = tanh(W_h h_t + b_h) \quad (13)$$
$$\alpha_t = \frac{\exp(e_t' c_h)}{\sum_{i=1}^{T} \exp(e_i' c_h)} \quad (14)$$
$$r = \sum_{t=1}^{T} \alpha_t h_t \quad (15)$$

The weight $\alpha_t$ represents the attention weight. It indicates the importance of the time-step $t$ observed value for prediction, which allows our model to concentrate or put attention on certain parts of the input time series for the final prediction. And we use *softmax* to normalize the vector $e_t$ of length $T$ to be the attention operation over the input time series sequence. $h_t$ represents the output hidden states of the GRU layer, and $r$ indicates the final shared representation with attention of each input sequence data.

## 4. Experiments

In this section, we use the real UK traffic flow datasets for the experiment to evaluate the proposed method. We also conduct comparative experiments on classic shallow models and baseline deep learning models to demonstrate the forecasting performance of the HMDLF framework.

### 4.1 Dataset

The real traffic datasets for experiments contain different attributes, e.g., location, date, time period, speed, flow, and journey time, etc. Details of the experimental dataset are described as follows:

Table 1. Experiments datasets description

| Dataset | Highways England |
|---|---|
| Datatype | Time series |
| Location | between M1 J7 and A405 |
| Intervals | 15-minute |
| Time Span | 01/01/2013-31/02/2014 |
| Attributes | flow, speed, journey time |
| Records | 37564 |

**Highways England dataset.** It's derived from Highways England Traffic Data from Opening up Government of UK [30]. This dataset provides average journey time, traffic speed and flow information for 15-minute periods on all motorways managed by the Highways Agency, known as the Strategic Road Network, in England. The dataset time span used for model training is from 01/01/2013 to 31/12/2013, which has 34876 records and another month (01/02/2014-31/02/2014, include 2688 records) data are used for testing.

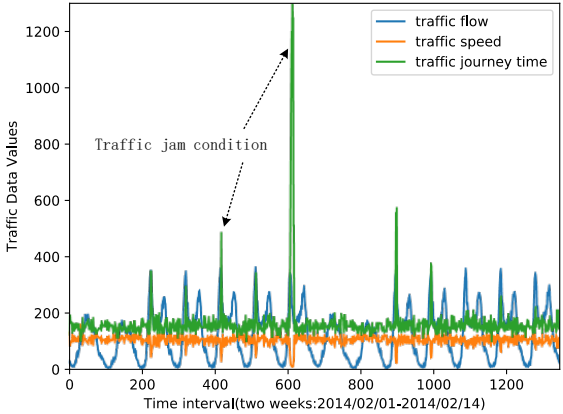

Fig. 6. A relative show of traffic flow, traffic speed and traffic journey time of two weeks data from test dataset (2014/02/01-2014/02/14).

A relative analysis of traffic flow, traffic speed and traffic journey time data (selected two weeks from test dataset) is illustrated in Fig. 6, which shows the nonlinear relationship of the three sequences data. From the data of traffic speed and traffic journey time, it can show whether the traffic congestion occurs or not since it is general that the traffic journey time is high and the traffic speed is low under the traffic jam condition.

*4.2 Experimental Setup*

Here we describe the experimental details and parameter settings of these models. The python libraries Keras which is based on Tensorflow is used to build our models. All experiments are performed by a PC Server (the configuration is Intel(R) Xeon(R) CPU E5-2623 3.00GHz, memory 128GB, 4 GPUs each is 12G NVIDIA Tesla K80C).

**Baselines.** The proposed framework is compared with several baseline traffic flow forecasting methods which are listed as follows.

SVR is a classic discriminative regression prediction method and the kernel-based SVR can make it possible to learn nonlinear trend in the training dataset. There are three kernels, which are SVR-RBF with RBF kernel, SVR-POLY with the poly kernel and SVR-LINEAR with the linear kernel.

ARIMA is one of the most widespread and prevalent models for time series prediction. LR represents linear regression model and DTR represents the decision tree regression model. RIDGE regression is also a statistical model which can alleviate multicollinearity problem amongst regression predictor variables.

RNN is a classic deep learning method for handling sequence learning tasks. GRU and LSTM are most popular variants of RNN. CNN is convolutional neural networks, which also can be used for time series modeling.

CNN-GRU and CNN-LSTM are the basic multimodal feature level learning modules of our proposed framework HDMLF and they are used for local trend features and long dependency features learning.

**Training.** The most difficult aspect of deep learning modeling is hyper parameter optimization. Deep neural networks need to set a lot of parameters. The default parameters in Keras are used for network initialization (such as weight initialization and learning rate). In order to avoid over-fitting of the deep neural network. We apply several common methods to solve this problem. A dropout with a probability of 0.2 is used in all fully connected layers and the batch size is set to 512. We use *tanh* as the activation function of the GRU and *ReLU* as the activation function of the CNN [28]. In addition, we use Adam as the optimizer, which has been shown that it has good generality and fast convergence ability in deep learning models. The baseline model's network structure (RNN, GRU, LSTM, CNN) uses one hidden layer and the number of neurons of each hidden layer is 128.

In the output of our proposed method (see Fig. 4), we use the linear function as the final activation function. Furthermore, we use min-max normalization method to scale each modality traffic data to [0, 1]. The HMDLF model is trained by optimizing the mean square error (MSE) loss function. Additionally, we select the traffic data in 2013 for training and validation (80% for training, and the rest 20% for validation), and select the January data in 2014 for testing. Through the cross-validation of the training set, we use early stopping strategy (and the patience parameter is 10) for training of deep neural networks. Furthermore, the penalty coefficient C of the SVR model is selected by grid search in training.

**Evaluation Metric**: Root Mean Square Error (RMSE) is used to evaluate the experimental results.

$$RMSE = \sqrt{\frac{1}{n}\sum_{i=1}^{n}(y_i - \hat{y}_i)^2} , \qquad (16)$$

where $\hat{y}_i$ and $y_i$ are the predicted value and ground truth, respectively, and $n$ is the number of all predicted values.

*4.3 Results*

The quantitative results are reported in Table II, which gives model error comparative analysis of ARIMA, SVR (different kernel), LR, DTR, RIDGE, RNN, LSTM, GRU, CNN, CNN-LSTM, CNN-GRU and our proposed framework HMDLF (with three different modules: CNN-LSTM, CNN-GRU and CNN-GRU with attention mechanism). It is found that

HMDLF achieves the best performance than the other methods in terms of prediction accuracy. Compared to the baseline models, our model HDMLF (with CNNGRU-Attention module) reduces error to 4.35, which significantly improves the accuracy. The RMSEs of baseline deep learning models are similar, especially CNN, CNN-LSTM and CNN-GRU. It implies that training single modal data cannot improve the performance obviously.

Table 2. RMSE of the proposed HMDLF model and comparisons with other baseline models for the traffic flow single-step forward forecasting task

| Type | Models | RMSE |
|---|---|---|
| shallow learning models for single modality (traffic flow) | SVR-POLY | 41.64 |
| | SVR-RBF | 15.41 |
| | SVR-LINEAR | 16.21 |
| | ARIMA | 14.01 |
| | RIDGE | 12.56 |
| | LR | 12.54 |
| | DTR | 12.51 |
| baseline deep learning models for single modality (traffic flow) | RNN | 12.18 |
| | CNN | 9.34 |
| | LSTM | 11.14 |
| | GRU | 11.15 |
| | CNN-LSTM | 9.75 |
| | CNN-GRU | 9.09 |
| our model for multi-modality(flow/speed/journeytime) | **HMDLF (with CNNLSTM module)** | **5.23** |
| | **HMDLF (with CNNGRU module)** | **4.67** |
| | **HMDLF (with CNNGRU-Attention module)** | **4.35** |

*Note: The baseline deep learning models and our model HMDLF parameters configuration are as follows: look-up size is 20, batch-size is 512 and dropout is 0.2. We used early stopping strategy for training and the patience parameter is 10.*

Moreover, the prediction performances of HMDLF (especially used CNN-GRU with attention mechanism model as the basic module has the best performance) and baseline deep learning methods are better than those of the classic shallow machine learning methods such as SVR and ARIMA. The reason is that HMDLF makes full use of local trend distribution, short-term temporal variability and long-term dependencies of single-modal traffic data. Furthermore, the proposed method also makes full use of the interdependence and exploiting this interdependence of multi-modality traffic flow related sequence data. Experimental results show that our hybrid multimodal deep learning framework can improve the performance by fusing those deep features information.

Through comparative analysis of experimental data, we also find that using GRU compared to LSTM will result in better predictive performance in our hybrid model. This is because GRU is simpler and have less hyper-parameters than LSTM and thus easier to modify, but still has the same performance as LSTM, in some times GRU will take much less time to train and even more efficient, and the use of CNN-GRU basic module in our hybrid model shows the best forecasting performance, which verified in our experiments that GRU is better than LSTM in some cases.

Then, we investigate the impact of epochs on different models. Fig. 7 shows the error curve of the proposed HMDLF model versus different epochs and comparisons with other baseline deep learning models. With the increase in epochs, the performance of all the models can be improved. This indicates that the RMSE slightly decreases in prediction when the epoch size is not very large. Especially, as the number of epochs increases, our model HMDLF (with CNN-GRU Attention module) always keeps the best performance over the other baseline models. And using CNN-GRU as the basic module has better performance than using CNN-LSTM.

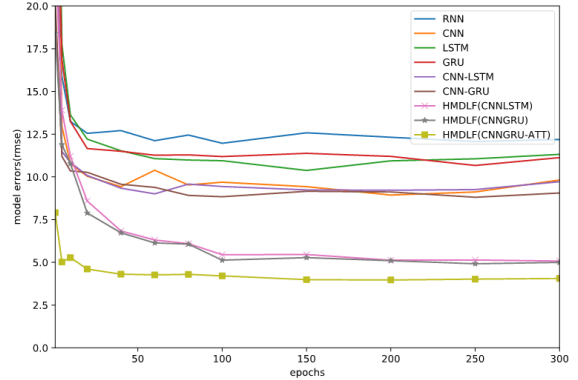

Fig. 7. RMSE of the proposed HMDLF model versus different epochs and comparisons with other baseline deep learning models.

Note that the RMSE achieves the lowest value when the epoch size is around 150 and remains almost steady when the epoch size continues to increase. That is to say, not the more epochs, the better the prediction performance is. The generalization capability cannot improve obviously when the epoch size is larger than 150. Moreover, all models seem to be little over fitting when the epoch size is larger than 250. In other words, the big number of epochs not only results in over fitting problems but also leads to great computational cost, although it can improve the accuracy of model training, which is not good for the application of the model.

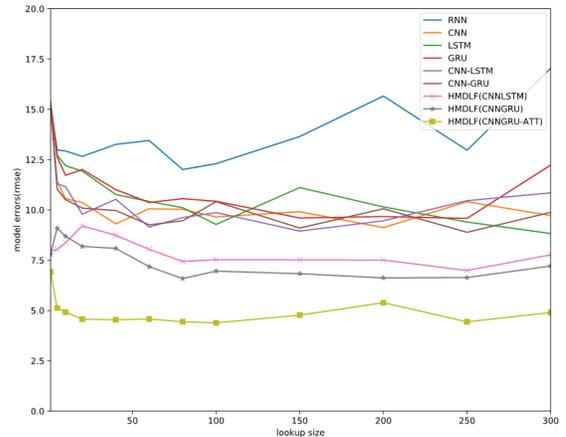

Fig. 8: RMSE of the proposed HMDLF model versus different lookup sizes and comparisons with another baseline deep learning models.

In addition, we analyze the impact of lookup size among different deep learning models. We use historical observations (the length is called window size or lookup size) to forecast the traffic flows in subsequent time intervals. From Fig. 8, we observe that compared to baseline deep models, our model HMDLF (with CNN-GRU Attention module) has the lowest error on the prediction across different lookup sizes. As the lookup size increases, the prediction errors of these baseline models decrease or back and forth oscillations such as RNN and LSTM. The RMSE of these contrast models reaches the minimum when the lookup size is between 50 and 100, and the RMSE remains stable or increases when the lookup size continues to increase because of over fitting problem. It is obvious that using CNN-GRU as the basic module has better prediction ability than using CNN-LSTM as the basic module by the comparison shown in Fig. 8.

To further evaluate the prediction performance of our proposed method, we examine the traffic flow prediction of HMDLF (with CNNGRU-Attention module) model over the course of one day (including 96 observed traffic flow data points) and three days (including 288 observed data points) traffic data of the testing dataset. We select two baseline models (SVR-RBF and LSTM) for contrast analysis with the proposed model.

Fig. 9 gives a comparison of the real (expected) traffic flow and predicted traffic flow values of two baseline models (the classic shallow machine learning model SVR with RBF kernel and the deep learning model LSTM) and our method HMDLF with CNN-GRU Attention module. The comparative analysis shows that the traditional deep learning model as LSTM, whose performance is better than that of shallow model SVR, and SVR cannot effectively predict the trough and peak values of traffic flow such as 9 am or 6 pm and so on. Although the prediction performance of LSTM is excellent, but it is not accurate enough to predict the peak and trough values of the traffic flow as our proposed method. The proposed HMDLF model can effectively predict the traffic peak condition with high accuracy in comparison to the ground truth data. Additionally, although the predictable performance of LSTM is better than SVR obviously, the gap between the predictable performance of LSTM and HMDLF on the figures is not significant due to the powerful learning ability of deep learning models (See Fig. 9, b and c).

Fig. 10 gives a comparison of the real (expected) traffic flow and predicted traffic flow values of SVR, LSTM and our model during three days (from 02/06/2014-02/08/2014, including two weekdays and one weekend day). The comparative analysis shows that SVR model cannot effectively predict the trough and peak traffic flow value, and the overall prediction performance of LSTM is better than SVR, either on weekdays or on weekends. LSTM for traffic flow forecasting can be carried out accurately not only at the trough values but also at the peak values of traffic time series data. The Fig. 10, (c) also shows that the overall prediction performance of our model (HMDLF with CNN-GRU Attention module) is excellent, and the traffic flow forecasting can be carried out accurately and effectively, not only during the general time period but also during the trough and peak time periods. In addition, because deep learning models have strong learning ability, the difference between our model and LTSM model for traffic flow prediction performance comparison is not obvious.

Finally, for better comparative analysis, we conduct a comparative experiment for model generalization ability (as shown in Fig. 11). The training data set and test data set of the above comparative experiment are from the same observation site (from Site A414 between M1 J7 and A405, named AL100). In the model generalization performance experiment, we also use AL100 training data set for model training, but use another obervation site's test data (named AL1811) set for testing. If the model is stable in traffic flow prediction at different observation stations, we can say that the model has better generalization ability.

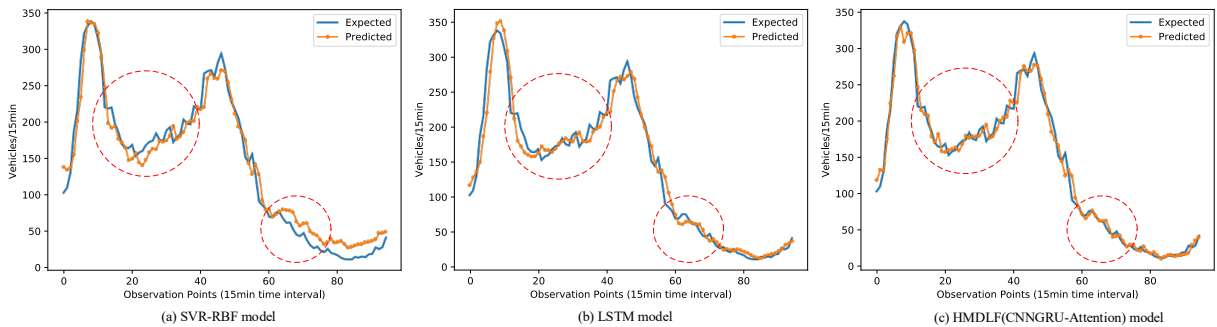

(a) SVR-RBF model　　(b) LSTM model　　(c) HMDLF(CNNGRU-Attention) model

Fig.9. Comparison of real observed (expected) and single-step forward predicted traffic flow of three models (SVR-RBF model, LSTM model and our HMDLF model with CNNGRU-Attention module) during one day (02/07/2014).

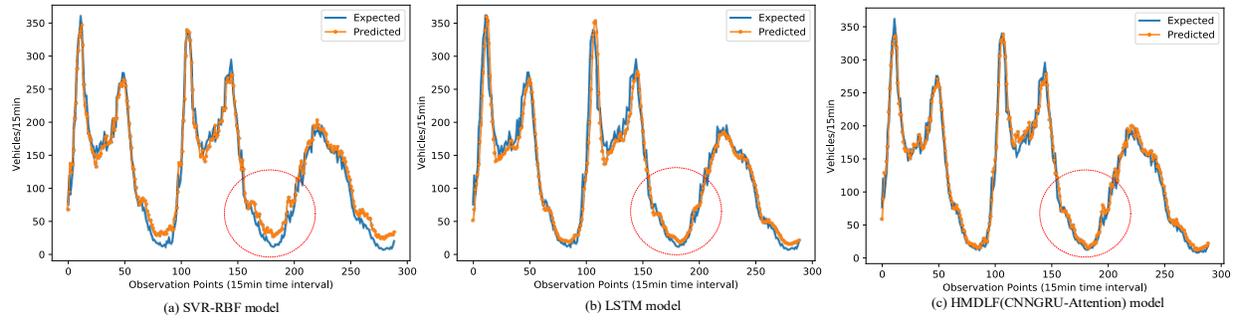

Figure 10. Comparison of real observed (expected) and single-step forward predicted traffic flow of three models (SVR-RBF model, LSTM model and our HMDLF model with CNNGRU-Attention module ) during three days ,include workdays and weekends(02/07/2014-02/09/2014).

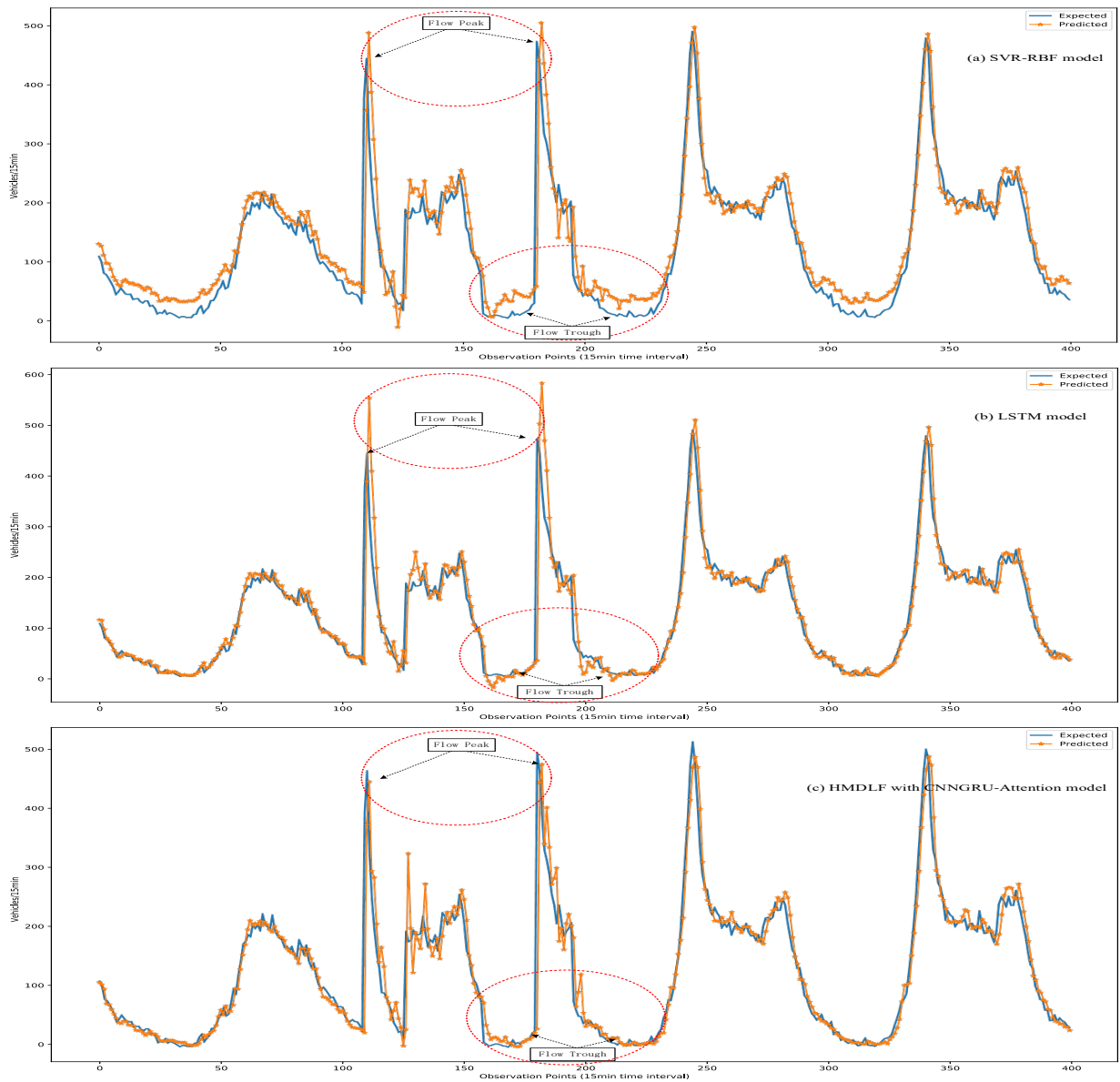

Fig. 11. Comparison of real observed (expected) and single-step forward predicted traffic flow of three models (SVR-RBF model, LSTM model and our HMDLF model with CNN-GRU Attention module) during 400 timesep points. And site AL100's data is used for model training, AL1811's data is used for prediction testing.

As the Fig. 11 shown, the comparative analysis indicates that the overall prediction performance and generalization ability of our model is the best, and the traffic flow forecasting can be carried out accurately and effectively, not only under the general condition but also under the congestion or accident conditions (at the trough or peak time periods). We can also see some interesting phenomena from the experimental comparison figures, especially in places with large prediction errors. The predicted value of the SVR model is higher than the ground-truth value, always at the traffic flow trough value. And the predicted value of LSTM model is higher than the ground-truth value at the flow peak, but the prediction value of the trough points is lower than the ground-truth value. And our model HMDLF is a good balance of this problem, which has the best comprehensive forecasting performance and generalization ability compared with shallow learning models and baseline deep learning models.

In summary, the comparative analysis of above figures shows that our model HMDLF is effective at peak or trough points forecasting with long traffic flow data (including weekdays or weekend, not only under normal condition but also under anomaly conditions, which as shown in the above figures). The traffic flow forecasting can be well matched with the expected reality, which means that our hybrid multimodal deep learning forecasting framework can effectively learn the trend, interdependence and spatial-temporal correlations of multimodal input traffic data. The multimodal deep learning structure of our traffic flow forecasting method can contribute to the development of intelligent transport systems.

## 5 Conclusion and Future Work

In this paper, we proposed an adaptively multimodal deep learning model HMDLF for short-term traffic flow forecasting. Firstly, our model integrates both one-dimensional CNN and GRU as one basical module to capture correlation features between local trends and long dependencies of single modality traffic data. Second, CNN-GRU auxiliary attention mechanism can more effectively explore and learn the deep nonlinear correlation features of multimodality input traffic data, e.g., traffic speed, flow, pass time and weather condition, etc., which is based on the jointly and adaptively multimodal representation and fusion learning framework by combined multiple CNN-GRU-Attention modules.

The major improvement in traffic flow forecasting of our method comes from multimodal learning by features fusion, which considers the correlation between different traffic flow data because traffic conditions are related to flow, speed, events, weather and so on. The main characteristics of our method include the local trends and long dependency learning of time series sequences, robust matching with error tolerance in peak and trough points, effective exploiting spatial-temporal interdependence of multimodality traffic data. The effectiveness of our model was verified by experiments on real traffic dataset from multiple angles, not only under weekdays and normal conditions but also under weekend and anomaly conditions.

As a future research direction, we believe that the traffic flow situation between adjacent traffic network nodes is interdependent. How to analyze and utilize this interdependence, which is important to improve the performance of traffic flow forecasting model, but effective data collection in short time period is a big obstacle. Moreover, the current experimental traffic data only have traffic flow, speed and pass time, due to the difficulty of collecting data on traffic accidents or extreme weather events, so the hybrid multimodal deep learning framework also needs to be further studied and improved by using more traffic datasets.